\pdfoutput=1
\documentclass[11pt]{article}

\usepackage[preprint]{acl}

\usepackage{times}
\usepackage{latexsym}

\usepackage[T1]{fontenc}
\usepackage[utf8]{inputenc}

\usepackage{microtype}

\usepackage{inconsolata}

\usepackage{graphicx}
\usepackage{color}
\usepackage{latexsym}
\usepackage{makecell}
\usepackage{amsmath,amssymb,bm}
\usepackage{url}
\usepackage {diagbox}
\usepackage[ruled, linesnumbered, titlenumbered]{algorithm2e}
\usepackage{bbm}
\usepackage{multirow}
\usepackage {booktabs}
\usepackage{pifont}
\usepackage{siunitx}
\usepackage{array}
\usepackage{colortbl}
\usepackage[utf8]{inputenc}
\usepackage{CJKutf8}
\usepackage{type1ec}
\usepackage[T2A,T1]{fontenc}
\usepackage[russian, portuges, ngerman, french, dutch, english]{babel}
\usepackage{textcomp}

\usepackage{caption}
\usepackage{footnote}
\makesavenoteenv{tabular}

\newcommand*{\ja}[1]{\begin{CJK}{UTF8}{ipxm}#1\end{CJK}}

\newcommand*{\red}{\textcolor{red}}

\title{Simultaneous Interpretation Corpus Construction\\ by Large Language Models in Distant Language Pair}

\author{
  Yusuke Sakai$^*$, Mana Makinae$^*$, Hidetaka Kamigaito, Taro Watanabe \\
  Nara Institute of Science and Technology \\
  \texttt{\{sakai.yusuke.sr9, makinae.mana.mh2, kamigaito.h, taro\}@is.naist.jp}
  }

\begin{document}
\maketitle

\def\thefootnote{*}\footnotetext{These authors contributed equally to this work.}\def\thefootnote{\arabic{footnote}}

\begin{abstract}

In Simultaneous Machine Translation (SiMT) systems, training with a simultaneous interpretation (SI) corpus is an effective method for achieving high-quality yet low-latency systems. However, it is very challenging to curate such a corpus due to limitations in the abilities of annotators, and hence, existing SI corpora are limited. Therefore, we propose a method to convert existing speech translation corpora into interpretation-style data, maintaining the original word order and preserving the entire source content using Large Language Models (LLM-SI-Corpus). We demonstrate that fine-tuning SiMT models in text-to-text and speech-to-text settings with the LLM-SI-Corpus reduces latencies while maintaining the same level of quality as the models trained with offline datasets. The LLM-SI-Corpus is available at \url{https://github.com/yusuke1997/LLM-SI-Corpus}.

\end{abstract}

\section{Introduction}

\begin{figure}[t]
\centering
\includegraphics[width=\linewidth]{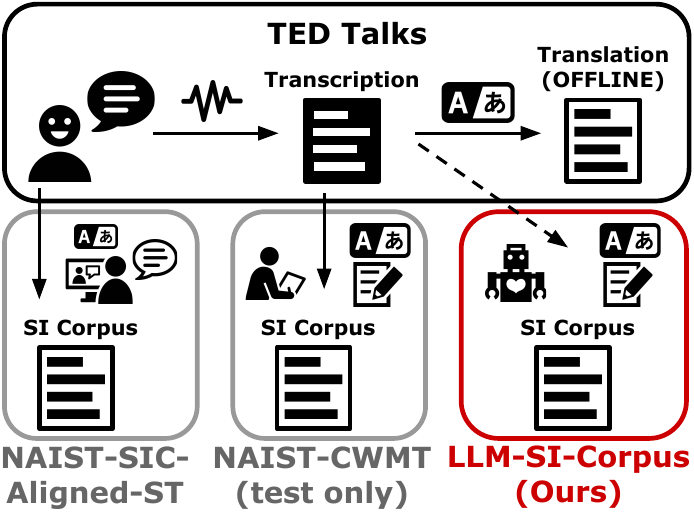}
\caption{The corpora used in this study, each created from the same TED Talks data. TED Talks are accompanied by English-Japanese offline MT data. NAIST-SIC-Aligned-ST~\cite{ko-etal-2023-tagged} is an SI dataset created by transcribing audio data of these talks by human interpreters. NAIST English-to-Japanese Chunk-wise Monotonic Translation Evaluation Dataset 2024 (NAIST-CWMT)~\cite{fukuda_chunk} is manually created based on offline MT data from TED Talks, following the CWMT guideline~\cite{okamura-2023}, and used only for testing purposes. Our LLM-SI-Corpus was created by LLMs based on the CWMT guideline and comprises training, development, and test sets.}
\label{fig:overview}
\end{figure} 

Simultaneous machine translation (SiMT)\footnote{Also, we called Simultaneous Speech Translation. We simplify the notation to SiMT in this paper for brevity.}~\cite{luong-manning-2015-stanford, gu-etal-2017-learning, ma-etal-2019-stacl, arivazhagan-etal-2019-monotonic} is a technique that translates input in real-time by incrementally processing partial segments rather than awaiting the whole sentence completion. While offline machine translation (MT) works without time restrictions, SiMT needs to start translating at certain points due to time limits, therefore balancing its latency and quality is crucial in this task. This becomes especially challenging with language pairs of drastically different word orders such as English and Japanese (SVO vs. SOV)~\cite{he-etal-2015-syntax, chen-etal-2021-improving-simultaneous, Deng_Ding_Liu_Zhang_Tao_Zhang_2023}. In the case of grammatically distant language pairs, one strategy involves maintaining the source language word order as much as possible to keep up with its original inputs to minimize its latency while maintaining its quality~\cite{cai-et-al-2020, han-etal-2021-monotonic, guo-etal-2023-simultaneous}. To address the balance between quality and latency, the best way to learn this interpretation strategy for SiMT systems is to utilize simultaneous interpretation (SI) data to train the model~\cite{ko-etal-2023-tagged}. While several SI datasets have been proposed for English and Japanese, they remain relatively limited in size compared to MT corpora. Furthermore, acquiring this data is costly and resource-intensive, making it impractical to scale it through manual dataset construction, resulting in a limited number of data. Therefore, current SiMT efforts focus on improving model performance and enhancing SI capabilities through data augmentation using a small amount of SI data.

Since such SI corpora are transcriptions of actual SI, there exist various transcription-related noises, e.g., fillers, and, therefore, it is still questionable whether they are optimal for training SiMT systems. Due to the specialized nature of SI, the quality of translation differs among interpreters due to the differences in the skills of individual interpreters and the varieties of their experience. Therefore, the quality of existing SI corpora is not consistent and includes SI techniques such as omissions and repetitions, making them not entirely faithful to the original speech and not ideal for training SiMT.

As a guideline of SI, \citet{okamura-2023} proposed Chunk-Wise Monotonic Translation (CWMT) for English to Japanese SI.
This guideline attempts to reduce latency by dividing speech into chunks and translating them by humans sequentially from the beginning. \citet{fukuda_chunk} manually created test data for SiMT 
%using the forward translation strategy 
in English to Japanese translation, confirming its higher fluency and validity compared to existing SI corpora. Their data construction guideline could potentially provide faithful translations with low latency and without omissions compared to existing SI corpora, but simply concatenating translated chunks at their boundaries results in unnatural translations. Furthermore, requiring human labor is an obstacle to scaling up the dataset size.

To solve these problems, we propose a method to convert existing speech translation (ST) corpora into SI-style data, maintaining the original word order and preserving the entire source content using Large Language Models (LLMs) as shown in Figure~\ref{fig:overview}. The LLM-SI-Corpus created by our method produces sentences that are closer to ideal SI compared to existing ST and SiMT corpora by faithfully following the CWMT guideline. These sentences feature reduced word order swapping, minimal omissions, and are more natural. We demonstrate that training SiMT models with the LLM-SI-Corpus text-to-text and speech-to-text settings improves translation quality in terms of semantics compared to existing SI corpora. Furthermore, the model trained with LLM-SI-Corpus reduces latencies while maintaining the same level of quality as models trained with offline datasets.

To summarize, our contributions are as follows:
\begin{itemize}
    \item We proposed a method for automatically constructing a training dataset for SiMT systems using LLMs focused on CWMT.
    \item We constructed the LLM-SI-Corpus, a large-scale training dataset for SiMT.
    \item We confirmed that the LLM-SI-Corpus, following CWMT, is effective for improving both quality and latency in SiMT systems.
\end{itemize}

\section{Background and Related Work}
\subsection{Simultaneous Machine Translation}
In the SiMT task, the model processes partial source sentences of length $J$ to incrementally generate partial target sentences of length $I$, guided by its policy.
Various policies have been proposed, primarily categorized as fixed and adaptive. 
Fixed policies~\cite{dalvi-etal-2018-incremental, ma-etal-2019-stacl, elbayad20_interspeech, zhang-feng-2021-universal} decide READ/WRITE operations based on predefined rules, such as the wait-$k$ policy~\cite{ma-etal-2019-stacl}, which reads $k$ source tokens initially and then alternates between writing and reading one token. 
Conversely, adaptive policies~\cite{zheng-etal-2020-simultaneous, liu20s_interspeech, papi-etal-2023-attention, Papi_2023} predict READ/WRITE operations based on the current source and target prefix, achieving a better balance between latency and translation quality.

\subsection{SI Corpora}

Most SI corpora are constructed from real-time human interpretation. In English to Japanese, several SI corpora are constructed~\cite{Tohyama-et-al-2004, shimizu-etal-2014-collection, doi-etal-2021-large}.
\citet{doi-etal-2021-large} developed a larger-scale SI corpus (NAIST-SIC) supporting both English to/from Japanese\footnote{They provide only a part of English-to-Japanese data.}. However, in NAIST-SIC, most data lack sentence alignment, rendering them unsuitable for model training.
To overcome this limitation, NAIST-SIC aligned text-to-text by \citet{zhao2024naistsicaligned} (NAIST-SIC-Aligned), speech-to-text by \citet{ko-etal-2023-tagged} (NAIST-SIC-Aligned-ST), hence it becomes a large-scale parallel English-Japanese SI corpus. \citet{fukuda_chunk} manually constructed a test dataset from NAIST-SIC-Aligned-ST based on chunk-wise monotonic translation strategy (described in Section~\ref{sec:cwmt}). For the other language pairs, \citet{pan-2019-chinese, zhang-etal-2021-bstc} (English-Chinese), \citet{kunz-etal-2021-heicic, zhao-etal-2021-good, machacek21_interspeech} (English-German), 
\citet{Paulik-and-Waibel-2009, bernardini2016epic, wang-etal-2021-voxpopuli, przybyl-etal-2022-epic} (the other language pairs include English) have been established.

However, SI corpus construction requires considerable time, money, and effort, resulting in a small corpus size. To address this challenge, \citet{he-etal-2015-syntax} attempted to mechanically reorder words in an MT corpus to convert it to an SI corpus by defining syntactic transformation rules for Japanese-to-English translation. However, spoken language poses challenges for syntactic parsing, and the rule-based approach tends to decrease fluency, making it difficult to apply the method to ST corpora.

\subsection{Chunk-Wise Monotonic Translation}
\label{sec:cwmt}
Chunk-wise monotonic translation (CWMT) and its variants are strategies employed by simultaneous interpreters, particularly in distant language pairs such as English and Japanese~\cite{Mizuno2016SimultaneousIA, okamura-2023,fukuda_chunk}.
This guideline addresses the grammatical disparities between the two languages, as preserving the exact word order from the source language to the target language can result in unnatural translations.
The necessity for this guideline arises from the need to strike a balance between translation latency and quality. Interpreters translating from English to Japanese prioritize maintaining the sequential order of information chunks from the source languages as much as possible.
In practice, interpreters divide sentences into manageable and meaningful chunks of information and translate them sequentially, ensuring that the order of chunks is maintained throughout the translation process. \citet{okamura-2023} defines these chunk boundaries and \citet{fukuda_chunk} defines its chunking workflow. The details of the guideline and workflow are described in Appendix~\ref{sec:workflow}.

\subsection{Style differences among SI, Offline Translation, and CWMT}

There are significant style gaps among SI, offline translation, and CWMT as described in \citet{fukuda_chunk, ko-etal-2023-tagged}. The examples are shown in Appendix~\ref{sec:style-difference-details}.
The findings include:
\begin{itemize}
    \item The SI translates the first half of the input earlier than the latter half, albeit with some unnaturalness and omission, whereas the offline translation preserves naturalness in Japanese through long-distance reordering from the input English (See Table~\ref{tab:offline-si_examples} in Appendix~\ref{sec:style-difference-details}).
    \item The offline translation and CWMT both include all content words from the source; however, their distinction lies in the order. In offline translation, long-distance reordering occurs to preserve naturalness, whereas, in CWMT, the order of source language chunks is maintained with some unnaturalness (See Table~\ref{tab:offline-cwmt_examples} in Appendix~\ref{sec:style-difference-details}).
\end{itemize}

From this observation, both SI and CWMT prioritize aligning source inputs as closely as possible, whereas offline allows for long-distance reordering. 
The significant difference in word order between English and Japanese poses a substantial challenge in SI, as highlighted in a prior study \cite{Mizuno2016SimultaneousIA}. 
Under the real SI scenario, interpreters prioritize delivering interpretation simultaneously to convey content promptly and preserve their working memory, which may involve some omission and summarization.
Consequently, our focus in this study lies in leveraging SI interpretation guidelines to train a SiMT model, aiming to enhance its simultaneity by prioritizing the preservation of word order in the source speech as much as possible.
The problem in CWMT lies in their approach to maintaining fluency, thus it is challenging to do automatically and it takes a high cost when annotating manually.

% ここからダイレクトに書いていく
\section{SI-Corpus Construction with LLMs}

To address the issues of the current SI corpus creation, we utilize LLMs, known for their high translation performance and the ability to perform purpose-specific translations based on given instructions~\cite{moslem-etal-2023-domain, zheng2024finetuning}. For our purpose, we follow the instructions for humans in creating CWMT to automatically convert existing ASR texts into SI corpora by LLMs.

\subsection{Prompt for Creating CWMT with LLMs}

\begin{figure}[t]
\centering
\includegraphics[width=\linewidth]{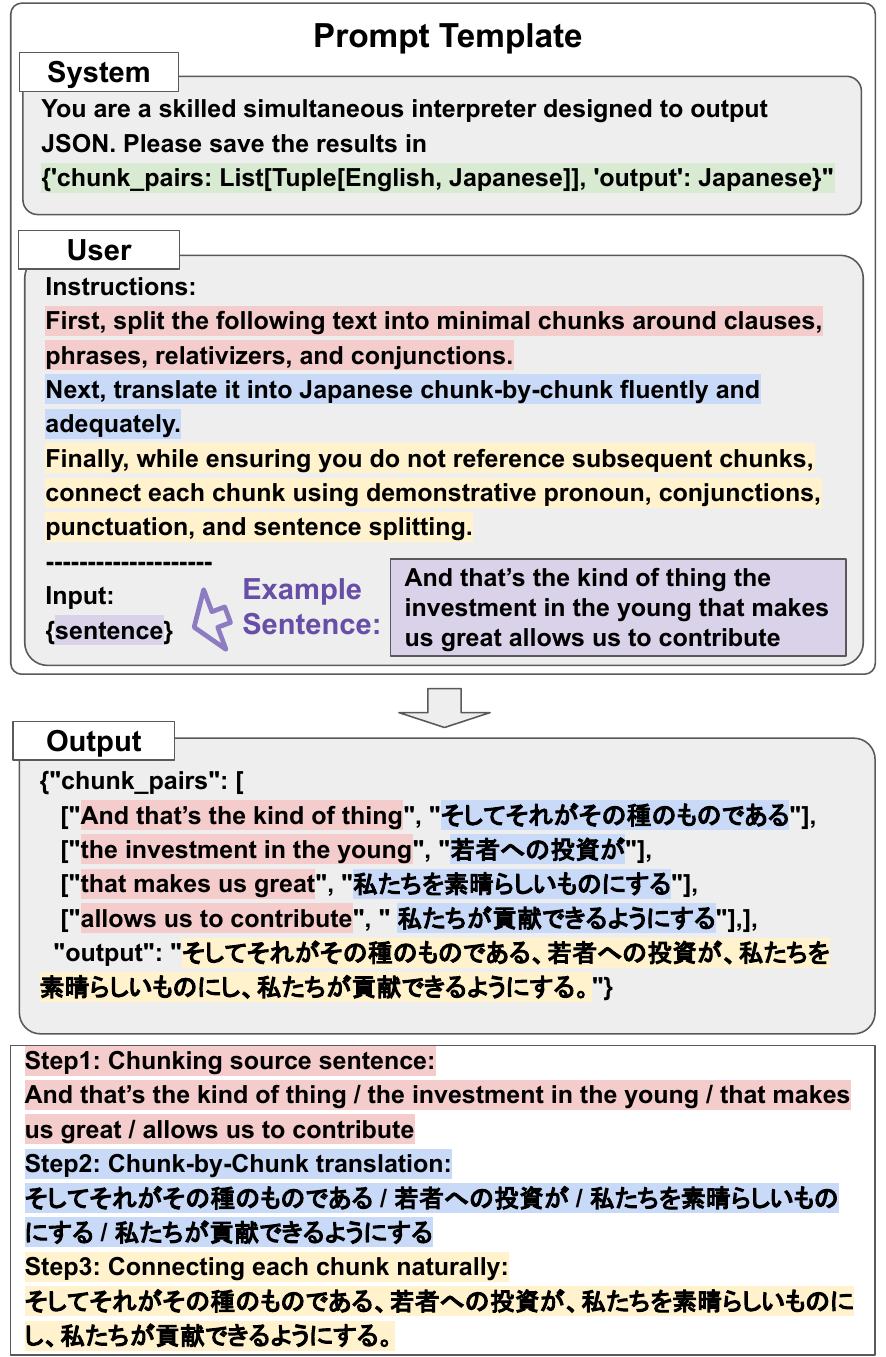}
\caption{The prompt template used for constructing the LLM-SI-Corpus. It is based on the CWMT workflow. This prompt can be divided into three color-coded steps.}
\label{fig:prompt}
\end{figure}

CWMT has three processes as described in Section~\ref{sec:cwmt}: chunking, translation of each chunk, and concatenating the translated chunks into sentences. Our prompt is based on the CWMT guideline by \citet{okamura-2023}, and we make it easy to understand for LLMs as described in Figure \ref{fig:prompt}. 

For chunking, we designed it to split around clauses, phrases, relativizers, and conjunctions. We dare to leave ambiguity in the instructions, making them more likely to be followed. 
Next, LLMs translate each chunk while maintaining fluency and adequacy.
Finally, LLMs generate CWMT by connecting chunks using only demonstrative pronouns, conjunctions, punctuation, and sentence splitting while ensuring not refer to subsequent chunks to preserve monotonicity. These processes are summarized in a single prompt\footnote{In the pilot study, we found similar results when we input data for each process separately as a pipeline or all at once into the LLMs. Thus, to address the cost issue, we chose to input all data at once as the prompt.}.
To ensure that these operations are performed according to the instruction without shortcuts, outputs are formatted in JSON\footnote{\url{https://platform.openai.com/docs/guides/text-generation/json-mode}} to obtain results at each step at once\footnote{We also employ various prompt tuning techniques, such as adding specific words to the instructions and using delimiters. Most of the prompt tuning techniques used in this study are described in \citet{bsharat2024principled}.}.

Note that there is no guarantee that LLMs will perform these operations correctly; however, the goal of the dataset construction is to improve the quality and latency of the SiMT system, so it is sufficient if the overall results predominantly include CWMT-like sentences\footnote{We use only the source side sentences to create CWMT, as using the target side proved to be noisy due to influence the noises of target side sentences according to our pilot study.}.

\subsection{Original Dataset Selection}

In this study, we focus on the English-Japanese direction and have selected the NAIST-SIC-Aligned-ST corpus~\cite{ko-etal-2023-tagged} as the original dataset. 

As shown in Figure 1, the NAIST-SIC-Aligned-ST corpus is based on TED Talks consisting of audio aligned with sentence-by-sentence transcriptions (offline translation) and manually created SI data, which allows direct comparison of the LLM SI-Corpus with manually created SI data\footnote{This type of dataset currently exists only in the NAIST-SIC dataset family~\cite{shimizu-etal-2014-collection, doi-etal-2021-large, zhao2024naistsicaligned, ko-etal-2023-tagged, fukuda_chunk}; thus, we are only testing the English-Japanese direction. Other language pairs are planned for future work.}. Furthermore, \citet{fukuda_chunk} have manually annotated CWMT transcriptions from the NAIST-SIC-Aligned-ST corpus, allowing us to compare CWMT qualities by humans and LLMs.

\subsection{LLM SI-Corpus Construction by LLMs}
We created two types of corpora with two different LLMs, GPT-3.5\footnote{gpt-3.5-turbo-0125}~\cite{gpt3-5} and GPT-4\footnote{gpt-4-0125-preview}~\cite{openai2024gpt4}. GPT-4 is known to have a higher ability to follow instructions and generate higher-quality outputs than GPT-3.5. Therefore, by comparing the two corpora, we also examine the differences in the abilities of the LLMs. The number of datasets in the LLM-SI-Corpus for train, dev, and test is 65,083, 165, and 511, respectively, which matches the numbers for NAIST-SIC-Align-ST. The total cost of data creation was 20 dollars (0.0003 dollars per sentence) for GPT-3.5 and 400 dollars (0.006 dollars per sentence) for GPT-4. 

\section{Experimental Setup}
\label{sec:experimental-setup}

To evaluate the effectiveness of the LLM SI-Corpus, we conduct experiments in Speech-to-Text settings. We also conducted Text-to-Text settings in Appendix~\ref{sec:text-to-text}, which shows a similar trend to Speech-to-Text.
We implemented the baseline using Fairseq~\cite{ott-etal-2019-fairseq, wang-etal-2020-fairseq} and SimulEval~\cite{ma-etal-2020-simuleval} and then applied test-time wait-$k$~\cite{ma-etal-2019-stacl} decoding policy to the offline models\footnote{We followed examples in GitHub repository: \url{https://github.com/ahclab/naist-simulst}}.

\paragraph{Speech-to-Text Settings}
Following the settings of~\citet{fukuda-etal-2023-naist, ko-etal-2023-tagged}, we employ pre-trained language models for both encoder and decoder by integrating them into the Transformer architecture~\cite{transformer}.
We used Hubert-Large~\cite{hubert} as the encoder, and we used the decoder parts of mBART50~\cite{tang-etal-2021-multilingual}, an encoder-decoder model pre-trained with 50 language pairs. We train the model with MuST-C v2.0~\cite{MuST-C} as continuous pre-training.
We fine-tuned the models for 3K steps, evaluating their performance every 200 steps, and terminated the fine-tuning if there was no improvement in the loss score for eight consecutive evaluations. The detailed settings are described in Appendix~\ref{sec:experimental-setup-details}.

\paragraph{Evaluation}

\begin{table}[t]
\centering
\small
\setlength{\tabcolsep}{4pt}
\begin{tabular}{@{}lcccc@{}}
\toprule
Quality Metrics & Textual & Meaning & Reference & Source \\
\midrule
BLEU    & \ding{51} &  & \ding{51} & \\ \midrule
BLEURT   & & \ding{51} & \ding{51} & \\
COMET    & & \ding{51} & \ding{51} & \ding{51} \\
COMET-QE  & & \ding{51} &  & \ding{51} \\
\bottomrule
\end{tabular}
\caption{Quality metrics used in our experiments}
\label{tab:compare-metrics}
\end{table}

Table~\ref{tab:compare-metrics} shows a list of translation quality evaluation metrics used in our experiments\footnote{We also evaluated with BERTScore~\cite{Zhang2020BERTScore}, but the trend is very similar to BLEURT.}. BLEU~\cite{post-2018-call} focuses on textual n-gram matching between the generated sentences and their reference sentences.
BLUERT~\cite{pu-etal-2021-learning}, COMET~\cite{rei-etal-2020-comet}, and COMET-QE~\cite{chimoto-bassett-2022-comet} utilize embeddings from language models to focus on semantic meanings.
BLUERT evaluates the generated sentences against reference sentences, while COMET also considers both source sentences and reference sentences.
In contrast, COMET-QE directly assesses the similarity between the source and generated sentences, thus avoiding the ambiguity that may arise from using references.
For latency evaluation metrics, we choose Average Lagging (AL)~\cite{ma-etal-2019-stacl}, Length Adaptive Average Lagging (LAAL)~\cite{papi-etal-2022-generation}, and Average Token Delay (ATD)~\cite{kano23_interspeech}\footnote{We cover all evaluation metrics used in the shared task of IWSLT 2024: \url{https://iwslt.org/2024/simultaneous}.}. We reported to test-time wait-$k$ in a set of $k$ is 1 to 35 at two intervals. One unit for $k$ was set to 160 frames for the speech-to-text setting. When $k$ = 3, after reading 3 $ \times$ 160 frames, the model would WRITE and READ alternately.

\paragraph{Datasets}
For training datasets, we utilize two versions of LLM SI-Corpus (GPT-4 and GPT-3.5) as proposed datasets, in addition to three baseline datasets: NAIST-SIC-Aligned-ST (SIC), offline data corresponding to SIC (OFFLINE), and a pre-trained model only (Pre-train).
For evaluation datasets, we compare using three types: the MuST-C tst-COMMON dataset (tst-COMMON), the test dataset from NAIST-SIC-Aligned-ST\footnote{\url{https://dsc-nlp.naist.jp/data/NAIST-SIC/Aligned-ST}} (SIC-test), and the chunk-wise evaluation dataset proposed by \citet{fukuda_chunk}, NAIST English-to-Japanese Chunk-wise Monotonic Translation Evaluation Dataset 2024\footnote{\url{https://dsc-nlp.naist.jp/data/NAIST-SIC/Aligned-Chunk_Mono-EJ}} (Chunk-wise).

\section{Experiments on Speech-to-Text Settings}
\label{sec:s2t-results}

\paragraph{Evaluation 1: tst-COMMON}

\begin{figure*}[!t]
\centering
\includegraphics[width=\linewidth]{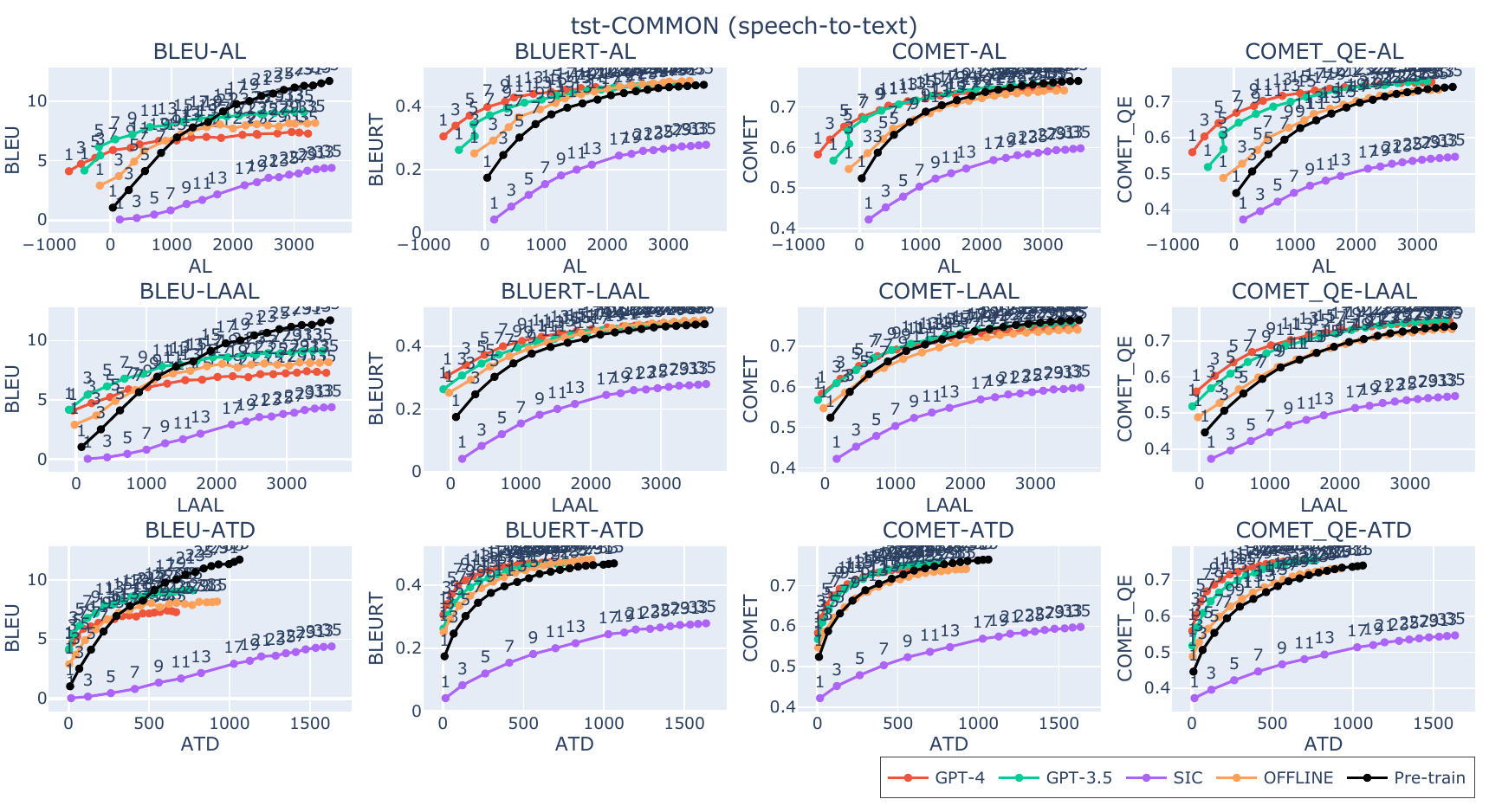}
\caption{The results of tst-COMMON dataset on speech-to-text settings. The values for each plot indicate $k$ of test-time wait-$k$.}
\label{fig:s2t-common}
\end{figure*}

Figure~\ref{fig:s2t-common} shows the results of speech-to-text experiments. 
When we focused on BLEU-AL in Figure~\ref{fig:s2t-common} for $k$ = 1, $k$ = 3, and $k$ = 5, the LLM SI-Corpus (GPT-3.5 and GPT-4) achieved higher BLEU scores than OFFLINE, indicating improvements in both latency and quality. 
However, as the value of $k$ increases, the BLEU score in Pre-train starts to surpass that of LLM SI-Corpus and OFFLINE when AL exceeds around $k$ = 9. 
This pattern persists across LAAL and ATD as well. 
This is attributed to the alignment of training and evaluation data, leading to enhanced BLEU scores.
Next, in \{BLEURT, COMET\}--\{AL, LAAL\}, both quality and latency in LLM SI-Corpus (GPT-3.5 and GPT-4) surpasses OFFLINE and Pre-train. 
Also in COMET-QE, the LLM SI-Corpus demonstrates superior quality and latency performance at all latencies in AL, LAAL, and ATD, indicating that the model trained on the LLM SI-Corpus can perform high-quality translations with low latency. 
Despite the trends observed in text-to-text settings, the quality gap remains evident in speech-to-speech settings even as $k$ increases.

\paragraph{Evaluation 2: SIC-test}

\begin{figure*}[!t]
\centering
\includegraphics[width=\linewidth]{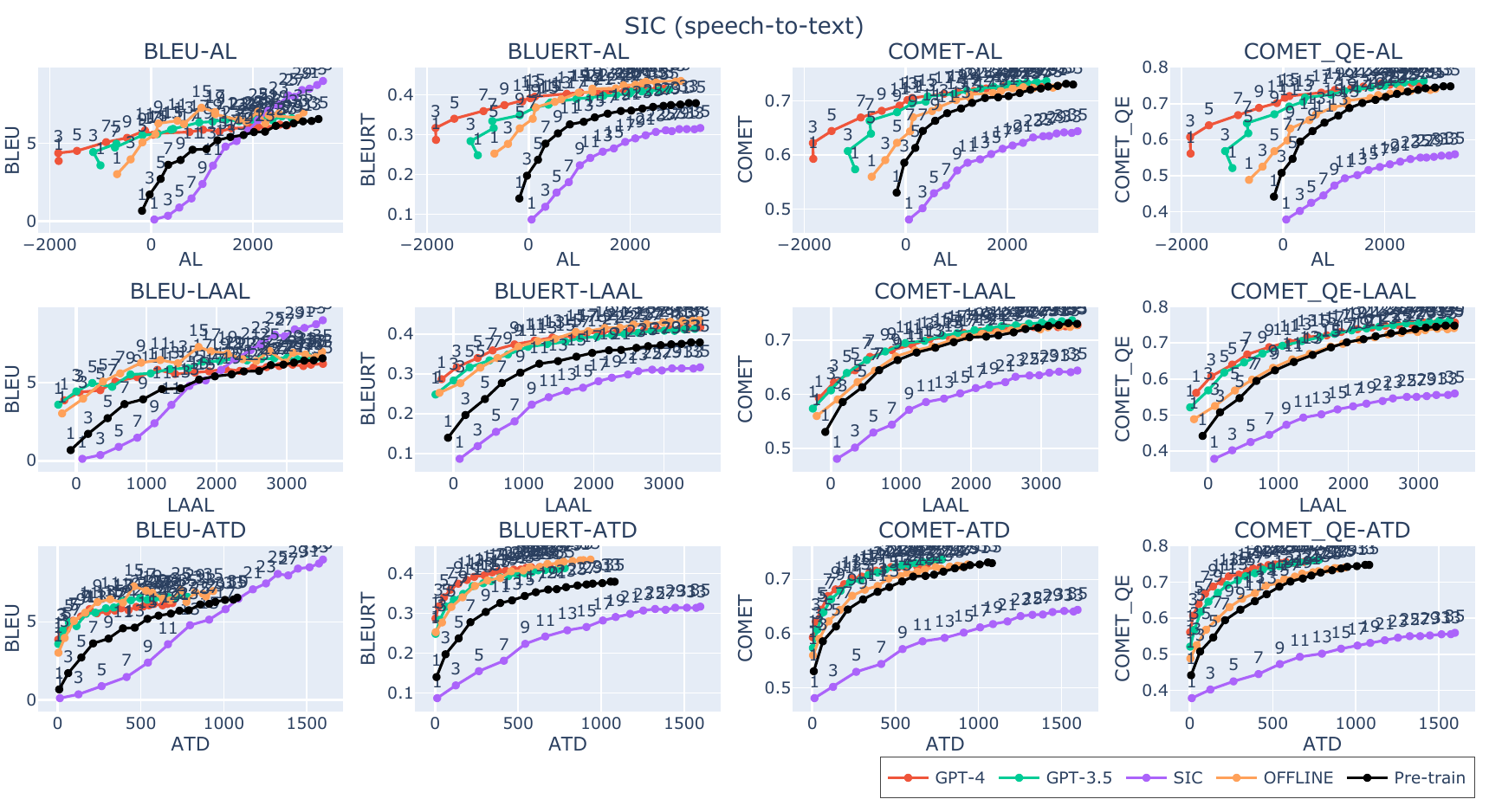}
\caption{The results of SIC-test dataset on speech-to-text settings. The notations are the same as Figure \ref{fig:s2t-common}.}
\label{fig:s2t-sic}
\end{figure*}

Figure~\ref{fig:s2t-sic} shows the result of SIC-test in speech-to-text settings. 
In Figure~\ref{fig:s2t-sic}, we focus on BLEU-AL, where the LLM SI-Corpus exhibits higher quality than OFFLINE up to around $k$ = 5. However, OFFLINE and SIC perform better at high latency because these align with the training and evaluation data, thereby improving the BLEU score.
The same trends are observed in LAAL and ATD.
Next, in \{BLEURT, COMET\}--\{AL, LAAL, ATD\}, both quality and latency in LLM SI-Corpus (GPT-3.5 and GPT-4) surpasses OFFLINE and Pre-train. 
In COMET-QE, the LLM SI-Corpus outperforms OFFLINE and Pre-train at all latencies in AL, LAAL, and ATD, indicating that the model trained on the LLM SI-Corpus can perform high-quality translations with low latency. 

\paragraph{Evaluation 3: Chunk-wise}

\begin{figure*}[!t]
\centering
\includegraphics[width=\linewidth]{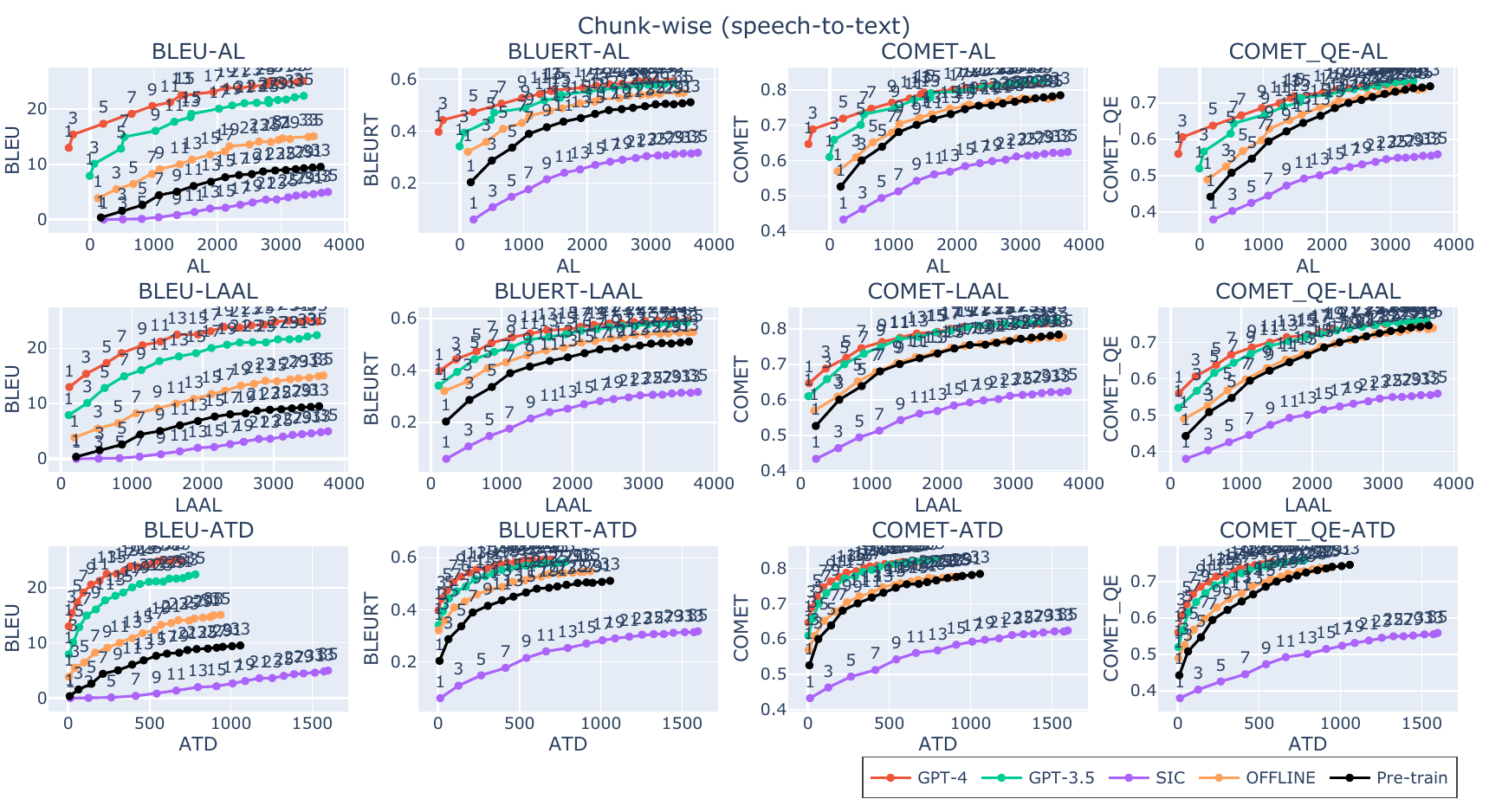}
\caption{The results of Chunk-wise dataset on speech-to-text settings. The notations are the same as Figure \ref{fig:s2t-common}.}
\label{fig:s2t-chunk-wise}
\end{figure*}

Figure~\ref{fig:s2t-chunk-wise} shows that the LLM SI-Corpus consistently exhibits superior quality and latency performance across all quality evaluation metrics. As previously observed in text-to-text, particularly when wait-$k$ is small, the quality gap among models is evident, albeit diminishing over larger wait-$k$ values. However, the gap remains noticeable in speech-to-speech settings.

\begin{table}[t]
\centering
%\resizebox{\linewidth}{!}{%
\footnotesize
\setlength{\tabcolsep}{3.5pt}
\begin{tabular}{@{}lrrrr@{}}
\toprule
\multicolumn{1}{c}{} & \multicolumn{4}{c}{Source: OFFLINE $\Rightarrow$ Target:}\\
 \cmidrule(){2-5} 
Metrics ($\uparrow$) & GPT-4 & GPT-3.5 & Chunk-wise & SIC-test  \\
\midrule
BLEU & 13.8 & 15.5 & 16.2 &  7.9 \\
ChrF & 26.2 & 25.8 & 28.5 & 16.3 \\ \midrule
BERTScore\_P & 77.9 & 78.1 & 78.4 & 75.2 \\
BERTScore\_R & 79.3 & 78.7 & 80.4 & 73.0 \\
BERTScore\_F1 & 78.6 & 78.3 & 79.3 & 74.0 \\ \midrule
BLEURT & 55.9 & 56.0 & 59.0 & 40.8  \\
COMET & 82.3 & 83.2 & 84.3 & 71.7 \\
COMET-QE & 82.6 & 82.8 & 82.9 & 63.1 \\ 
\bottomrule
\end{tabular}
%}
\caption{Similarities between OFFLINE and each SI corpus on the test set. BLEU and CharF indicate the similarities of textual alignment. BERTScore compares the semantic precision and recall between the source and target sentences. BLEURT, COMET, and COMET-QE compare semantic similarity, as shown in Table~\ref{tab:compare-metrics}.}
\label{tab:compare-offline-si}
\end{table}

\paragraph{Summary}
We evaluated the fine-tuned models with LLM SI-Corpus in three different test data.
The results indicate that the LLM SI-Corpus delivers the best translation quality with minimal latencies across all semantic similarity-focused evaluation metrics.
Even in BLUE, the LLM SI-Corpus achieves equivalent translation quality especially when $k$ is small. 

Meanwhile, in the SIC fine-tuned model on the ATD evaluation setting, we observed significantly longer lags compared to other fine-tuned models. This trend is also evident in \citet{ko-etal-2023-tagged} in Figure~\ref{fig:s2t-common}. This observation may be attributed to the fact that some transcripts in SIC are extremely short relative to the length of the source speech input. Fine-tuning with such data may lead to undesired generation results, such as excessive repetition (Table~\ref{tab:not-good-sic_examples} in Appendix~\ref{sec:discussion-details}), leading to longer lags.
It can be argued that while achieving a shorter output length is deemed advantageous in the ATD setting, the current evaluation system may disproportionately prioritize the shortest length, which may be considered unfair. Output results in excessive lengthening or shortening should be subject to penalties. We leave this as future work.

\section{Discussions}
We picked some important discussion themes. The more discussions are described in Appendix~\ref{sec:discussion-details}.

\subsection{Similarity between OFFLINE and SI}

\begin{table*}[!t]
\small
\centering
\begin{tabular}{cp{13cm}}
% \hline \textbf{Transcript} & \textbf{Example} \\
\toprule
Source & {\small (1) A few weeks later, / (2) the department / \underline{(3) received a letter} / \underline{(4) from the homeowner} / (5) thanking us / (6) for the valiant effort displayed in saving her home.} \\
\midrule 
Reference & \ja{\small (1) 数週間後 / (2) 消防団は / \underline{(4) 家主から} / (6) 火事の際の勇敢な活動に対する / (5) お礼の/ \underline{(3) 手紙をもらいました}。} \\
\midrule
Pre-train & \ja{\small (1) 数週間後 / (2) 政府は / \underline{(3) 手紙を送りました)}。} \\
\midrule
NAIST-SIC & \ja{\small (1) 数週間後、} \\
\midrule
Offline & \ja{\small (1) 数週間後、/ (2) 政府は、/ \underline{(3) 手紙を送りました}。} \\
\midrule
GPT-3.5 & \ja{\small (1) 数週間後、/ (2) その部門は / \underline{(3) 手紙を受け取った}。/ \underline{(4) 自宅のオーナーから}、/ (5) 私たちに感謝 / (6) の手紙を、安全を確保するために彼女の家を救うために示された勇敢な努力に感謝する。} \\
\midrule
GPT-4 & \ja{{\small (1) 数週間後、/ (2) その部門が/ \underline{(3) 手紙を} / \underline{(4) 自宅から所有者から} / \underline{(3) 受け取った}。/ (5) それは、私たちに感謝の意を表すもので、/ (6) 彼女の家を救うために勇敢な努力がなされた。}} \\
\bottomrule
\end{tabular}
\caption{Examples of the generated texts for $k$ = 7 in speech-to-text settings on tst-COMMON. The bracketed numbers indicate the corresponding phrases in the source text.}
\label{tab:simul_examples_which_better}
\end{table*}

Table~\ref{tab:compare-offline-si} shows a comparison of similarities between OFFLINE and SI in the test set. The results in Table~\ref{tab:compare-offline-si} indicate that Chunk-wise and OFFLINE are the most similar to OFFLINE across all evaluation metrics. Additionally, the SIC-test has significantly lower similarity compared to other SI corpora. Focusing on BERTScore's recall, unlike other CWMT-based SI corpora which show higher precision, it suggests the inclusion of omissions or dropped translations due to its lower precision. Furthermore, a comparison in COMET-QE between LLM-SI-Corpus and Chunk-wise shows equivalent quality, suggesting that the LLMs have capabilities comparable to manually created data in CWMT.

\subsection{Is the CWMT guideline effective for SI?}
% ここを修正中
It might be too early to conclude regarding the effectiveness of CWMT due to its limited availability only on test data. Meanwhile, access to CWMT-applied training data is unrealistic as it involves the interpreters' engagement in its creation. Based on our current observations of its test data, CWMT specializes in achieving complete synchronization for word order without any omissions. 
This characteristic aligns well with machine translation evaluation metrics, ensuring precise content correspondence between the source and target texts.
However, it overlooks the importance of summarization, which is also a key tactic utilized in SI scenarios for minimizing latency. Furthermore, its excessive focus on aligning word order in source speech results in unnatural translations.
Thus, creating an SI Corpus that considers factors such as omission is a necessary future challenge.

\subsection{Which is better GPT-4 vs. GPT-3.5?}
%修正完了
In terms of preserving word order, both versions of GPT demonstrate equivalent proficiency. This suggests a comparable ability to understand its prompts. If the primary objective is solely the preservation of word order, GPT-3.5 suffices. However, for those who prioritize output quality, GPT-4 may offer better performance as shown in Table~\ref{tab:simul_examples_which_better}. As demonstrated, both GPT-3.5 and GPT-4 maintain the word order in the source language. However, in some cases, reordering occurs in GPT-4. This type of reordering should be allowed as it enhances naturalness. Conversely, in GPT-3.5, it lacks fluency although the order is maintained. The details are described in Appendix~\ref{sec:qualitative-details}.

Meanwhile, in the SiMT results in Section~\ref{sec:s2t-results}, the fact that GPT-3.5 surpasses GPT-4 in some BLEU scores indicates that metrics considering only textual similarity cannot capture such an issue, necessitating the creation of new evaluation metrics.

Furthermore, the performance improvements over OFFLINE observed with both GPT-3.5 and GPT-4 suggest that LLMs with a certain level of instruction-following capability are effective for corpus construction.

\section{Conclusion and Future Directions}
% 追記完了
In this study, we proposed a method for converting ST corpora to SI corpora using LLMs, based on the CWMT guideline, and we constructed the LLM-SI-Corpus.
To verify the effectiveness of the LLM-SI-Corpus, we conducted experiments in three scenarios: a general offline ST corpus (tst-COMMON), an SI corpus (SIC-test), and a CWMT test corpus (Chunk-wise), in both speech-to-text and text-to-text settings. In all scenarios, the SiMT models trained with the LLM-SI-Corpus outperformed others with low latency and high quality.
Moreover, while manually constructing SI corpora is costly, the LLM-SI-Corpus can be produced for only 20 dollars. Therefore, it can be easily applied to other ST corpora or adapted to other languages since it utilizes LLMs.

As future directions, we plan to explore the application of other SI techniques such as omission, apply them to other large-scale ST corpora, and also expand their application to speech-to-speech settings.

%\clearpage

\section{Limitations}

\paragraph{Lack of SiMT evaluation data, methods, and definitions}

The existing metrics for evaluating SiMT systems have become a challenge in reducing latency for ST test data such as tst-COMMON, although SI involves various techniques, e.g., omission. Therefore, the use of ST data for evaluation is a major limitation of this work. Thus, there is an urgent need to establish evaluation metrics and data suitable for SiMT. Moreover, despite the various SI techniques available, there has been no mature discussion from an engineering perspective on which SI techniques are necessary and essential for SiMT; therefore, we need to address this issue in our future work. These issues were clarified through our comprehensive experiments and analysis.

\paragraph{Construct more SI corpora}
In this study, we constructed the LLM-SI-Corpus based on the NAIST-SI-Aligned-ST corpus for comparison with existing SI corpora. Our method is applicable to various other ST corpora at low cost. Moreover, this study has shown that the output of LLMs is effective for developing SiMT corpora; therefore, we plan to explore its applicability to other SiMT methods like omissions as our future steps. We hope that by expanding into multiple languages and enhancing data augmentation, further development in the SiMT field will be achieved.

\paragraph{Dataset Quality}
In this study, we used GPT-3.5 and GPT-4 with a simple prompt for data creation. Therefore, there is room for improvement in the selection of LLMs and the refinement of prompts. Thus, it may become possible to create higher quality datasets at a lower cost when the API prices decrease or by switching to other strong LMs such as Gemini~\cite{geminiteam2024gemini}, Claude 3 and Qwen~\cite{bai2023qwen}. Additionally, employing prompt strategies that leverage the capabilities of LMs, such as Chain of Thought (CoT)~\cite{wei2022chain}, Tree of Thought (ToT)~\cite{yao2023tree} and ReAct~\cite{yao2023react}, could potentially lead to the production of higher quality datasets.

\paragraph{Other SI techniques}
In this study, we addressed CWMT, specifically focusing on chunking within SI techniques. However, there are many other SI techniques~\cite{camayd2011cognitive, okamura-2023}, such as omission and summarization, and addressing these is also necessary to achieve better SI. Furthermore, the means of evaluating these techniques and systematic methods are still in development and not yet established, making this a key topic of focus for the SiMT field in the future.
However, LLMs have a certain understanding of the CWMT guidelines, which allows them to account for word reordering from the insight of this study. Therefore, the next step is to evaluate whether LLMs can comprehend omissions, and whether they can perform balanced omission and summarization based on the number of syllables. This will be explored as a future challenge.

\section{Ethical Considerations}

\paragraph{License of Source Dataset}

The NAIST-SIC-Aligned-ST corpus used in this study is available only for research purposes. We have used this corpus for research, so there are no license violations. Moreover, the LLM SI-Corpus was created from the NAIST-SIC-Aligned-ST corpus and thus inherits its terms of use\footnote{\url{https://dsc-nlp.naist.jp/data/NAIST-SIC/Aligned-ST/}}. In terms of distribution, redistribution of interpretation transcripts is prohibited; therefore, we release only our transcripts and the corresponding audio segment information and do not contain any audio data or the original transcripts. Furthermore, the README file of the LLM SI-Corpus clearly states the source of the data, the license, and acknowledgments, and properly documents the original data information. Note that, it is permitted to cite example sentences from the NAIST-SIC-Aligned-ST corpus.

\paragraph{Ownership rights about outputs of the LLMs}
The LLM SI-Corpus was created using GPT-3.5 and GPT-4 and is therefore subject to OpenAI’s license terms\footnote{\url{https://openai.com/policies/terms-of-use}}. OpenAI assigns to us all rights, titles, and interests in and to the output. As a result, we retain the ownership rights. There are no restrictions on distributing the datasets, but in line with NAIST-SIC-Aligned-ST, we distribute only for research purposes. However, these terms may change, and there may be a need to impose distribution restrictions depending on the terms.

\paragraph{Moderations}
Since the LLM SI-Corpus fundamentally originates from TED Talks, it does not contain any potentially harmful information. Furthermore, we checked using OpenAI Moderation APIs\footnote{\url{https://platform.openai.com/docs/guides/moderation}} and found no examples of harmful content.

\bibliography{anthology,custom}
%\bibliography{custom}

\appendix

\section{Detail of the CWMT Guideline and Workflow}
\label{sec:workflow}
%ここから下の記述については、最悪appendix飛ばしにしてもいいかもしれない。
%順送り訳のチャンキングのやり方
\citet{okamura-2023} defines these chunk boundaries using the following rules (rule 1, 2, 3, and 4), then \citet{fukuda_chunk} added the fifth rule as follows:
\begin{enumerate}
    \item Before conjunctions or relative pronouns that introduce clauses (excluding when they modify the subject).
    \item  After infinitives, prepositions, or gerunds when followed by three or more words.
    \item When the subject consists of three or more words.
    \item Before and after punctuation marks such as commas (excluding lists of individual words), semicolons, hyphens, etc.
    \item Before prepositional phrases or adverbial phrases following the beginning of a sentence (or directly after conjunctions or relative pronouns that introduce clauses).
\end{enumerate}

Based on these guidelines, \citet{fukuda_chunk} defines its chunking workflow.
First, rules 1, 3, 4, and 5 are applied to each source sentence chunk, and then the translated chunks are concatenated while preserving boundaries. Rule 2 is optionally applied in the last step to avoid the influence of the prior steps causing extremely small chunk translations.
This chunk-wise approach enables interpreters to navigate the challenges posed by grammatical differences between the source and target languages while managing the demands for translation speed and accuracy.

Based on this chunking workflow, \citet{fukuda_chunk} manually constructed a test dataset. The procedure is as follows:
\begin{enumerate}
    \item Translate each chunk from the beginning of the sentence.
    \item Translate in a way that the connection between chunks is natural when considering the entire sentence.
    \item Translate without including information from the following chunks. 
    \item Additionally, for the sake of maintaining the fluency of the sentence, the following operations are permitted, but applied carefully:
    \begin{enumerate}
        \item Repeating the information from the previous chunk.
        \item Deferring the information to be translated to the following chunk.
        \item Omitting unnecessary information.
    \end{enumerate}
\end{enumerate}

\section{Style differences among SI, Offline Translation and CWMT (Details)}
\label{sec:style-difference-details}

\begin{table*}
\centering
\small
\begin{tabular}{cp{13cm}}
% \hline \textbf{Transcript} & \textbf{Example} \\
\toprule
Source & And (1) I’m  / (2) not here to / (3) say that / (4) men are to / (5) blame for the / (6) crisis and what / (7) happened in my / (8) country. \\
\midrule
OFFLINE  & \ja{しかしこの経済 / (6) 危機や私の / (8) 国での / (7)  出来事について / (1) 私は / (4) 男性に / (5) 非があると / (3) 言うつもりは / (2) ありません} \\
\midrule
SI & \ja{(4)男性の、/ (5) せいだけでは / (2) ありません、私どもの / (8) 国の、金融 / (6) 崩壊の、/ (5) 責任は、}\\
\bottomrule
\end{tabular}
\caption{Translation style difference between offline and SI. The number indicates the corresponding words in the source. The example is coming from \cite{ko-etal-2023-tagged}.}
\label{tab:offline-si_examples}
\end{table*}

\begin{table*}
\centering
\small
\begin{tabular}{cp{13cm}}
% \hline \textbf{Transcript} & \textbf{Example} \\
\toprule
Source & (1) Groups like Anonymous / (2) have risen up / (3) over the last 12 months / (4) and have become a major player / (5) in the field of online attacks. \\
\midrule
OFFLINE  & \ja{(1) Anonymous というグループは / (3) この 12 ヶ月ほど / (2) 活気づいていて / (5) オンライン攻撃において / (4) 大きな 存在になってます。} \\
\midrule
CWMT & \ja{(1) アノニマスのようなグループが / (2) 台頭してきています， / (3) 過去 12 ヶ月にわたって， / (4) そして主要なプレイ ヤーになっています， / (5) オンライン攻撃の分野において.}\\
\bottomrule
\end{tabular}
\caption{Translation style difference between offline and CWMT. The number indicates the corresponding words in the source. The example is coming from \cite{fukuda_chunk}.}
\label{tab:offline-cwmt_examples}
\end{table*}

There are significant style gaps among SI, offline translation, and CWMT as described in \citet{fukuda_chunk, ko-etal-2023-tagged}.
Table~\ref{tab:offline-si_examples} and Table~\ref{tab:offline-cwmt_examples} are examples describing their differences.

\section{Experiments (Details)}
\label{sec:text-to-text}

\subsection{Experimental Setup}
\label{sec:experimental-setup-details}
The evaluation methods and datasets are the same as those described in Section~\ref{sec:experimental-setup}.

\paragraph{Speech-to-Text Settings}
Following the settings of~\citet{fukuda-etal-2023-naist, ko-etal-2023-tagged}, we employ pre-trained language models for both encoder and decoder\footnote{Our baselines are almost the same as the baseline of IWSLT2023 Speech-to-Text settings (\url{https://github.com/facebookresearch/fairseq/tree/iwslt2023/examples/simultaneous_translation}), but, due to an implementation issue, we have switched the encoder from wav2vec 2.0~\cite{wav2vec2.0} to HuBERT~\cite{hubert}.} by integrating them into the Transformer architecture~\cite{transformer}.% by Inter-connection~\cite{nishikawa23_interspeech}.
We used Hubert-Large~\cite{hubert} as the encoder, which includes a feature extractor and transformer encoder layers. The feature extractor, trained on 60k hours of unlabeled speech data from Libri-Light~\cite{librilight}, consists of a 7-layer convolutional network with kernel sizes of (10,3,3,3,3,2,2), strides of (5,2,2,2,2,2,2), and 512 channels.
For the decoder side, we use the decoder parts of mBART50~\cite{tang-etal-2021-multilingual}, an encoder-decoder model pre-trained with 50 language pairs. The decoder consists of 12 layers of transformer decoders, and the embedding layer and linear projection weights are shared, with a vocabulary size of 250K. The inputs are waveforms with a 16kHz sampling rate that are normalized to zero mean and unit variance. During training, each source audio is augmented~\cite{Eugene} with a probability of 0.8. We train the model with MuST-C v2.0~\cite{MuST-C} as continuous pre-training.
We fine-tuned the models for 3K steps, evaluating their performance every 200 steps, and terminated the fine-tuning if there was no improvement in the loss score for eight consecutive evaluations.
To avoid overfitting to the small SI data, the following parameters are fixed~\cite{tsiamas-etal-2022-pretrained}: the feature extractor and feed-forward layers of the encoder and the embedding, self-attention, and feed-forward layers of the decoder.

\paragraph{Text-to-Text Settings}
We train an NMT model through pre-training\footnote{Our baselines are based on the English-to-Japanese Text-to-Text translation at IWSLT2022 settings: \url{https://github.com/ksudoh/IWSLT2022_simul_t2t_baseline_enja}}, then fine-tuned it using SI data. For pre-training, we used WMT21 En-Ja datasets~\cite{akhbardeh-etal-2021-findings} (JParaCrawl v3~\cite{morishita-etal-2022-jparacrawl}, News Commentary v16~\cite{tiedemann-2012-parallel}, WikiTitles v3~\cite{tiedemann-2012-parallel}, WikiMatrix v1~\cite{schwenk-etal-2021-wikimatrix}, JESC~\cite{pryzant-etal-2018-jesc}, KFTT~\cite{neubig11kftt}) and MuST-C v2.0~\cite{MuST-C}.
We use SentencePiece~\cite{kudo-richardson-2018-sentencepiece} for subword tokenization with a Unigram Language Model~\cite{kudo-2018-subword}. The vocabulary size is 32K tokens with a character coverage of 0.99995 on a shared dictionary. The tokenizer was trained on the pre-training data. 
We use a Transformer-big model~\cite{transformer}, warmup update at 4000, dropout at 0.3, and the learning rate at 0.0005. 
The model is trained for 100K steps, with evaluation conducted every 2K steps.
Training is terminated if there is no improvement in the best loss after eight consecutive evaluations.
During fine-tuning, we trained for 3K steps, with evaluations conducted every 200 steps.
Fine-tuning is also finished if there are no updates after eight consecutive evaluations.

\subsection{Results on Text-to-Text Setting}

\begin{figure*}[!t]
\centering
\includegraphics[width=\linewidth]{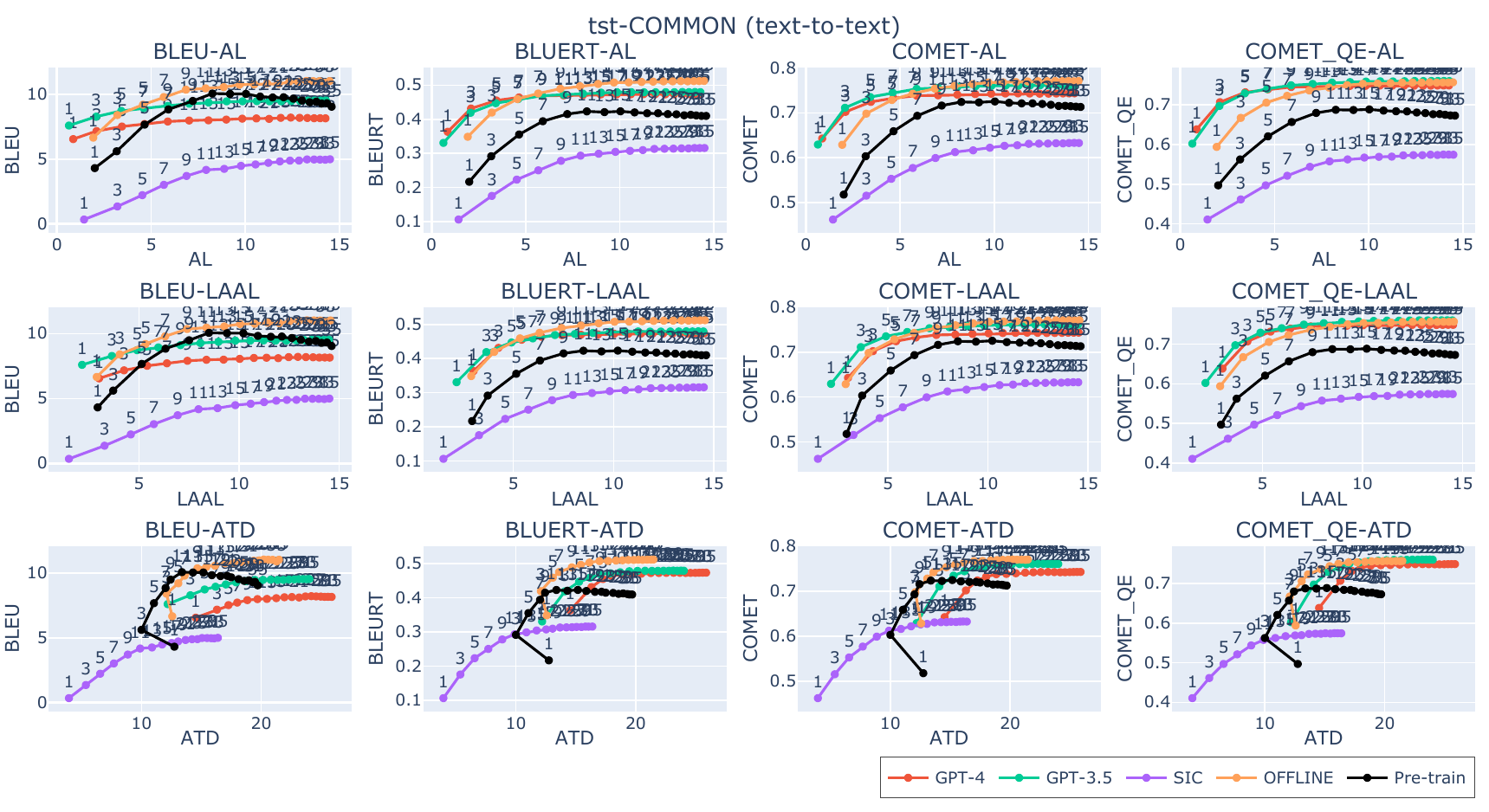}
\caption{The results of tst-COMMON dataset on text-to-text settings. The notations are the same as Figure \ref{fig:s2t-common}.}
\label{fig:t2t-common}
\end{figure*}

\paragraph{Evaluation 1: tst-COMMON}
Figure~\ref{fig:t2t-common} shows the result of tst-COMMON in text-to-text settings.
We focused on BLEU-AL in Figure~\ref{fig:t2t-common}, for $k$=1 and $k$=3, the LLM SI-Corpus (GPT-3.5 and GPT-4) achieves higher BLEU scores than OFFLINE, indicating improvements in both latency and quality. However, as the value of $k$ increases, the BLEU score begins to decrease compared to Pre-train when AL is more than 5. Similar trends are observed for LAAL.
Next, in \{BLEURT, COMET\}--\{AL, LAAL\}, the quality surpasses OFFLINE when the latency is approximately less than 5. Moreover, when compared with Pre-train, the translation qualities are improved at all latencies. 
In COMET-QE, which focuses on the semantic similarity between the source text and the generated text, the LLM SI-Corpus outperforms OFFLINE at all latencies, indicating that the model trained on the LLM SI-Corpus can perform high-quality translations with low latency. Conversely, in ATD, although the quality remains unchanged, an increase in latency is observed, suggesting that the output sequence length is longer compared to the source text. Meanwhile, all results of the SIC-test show a decline in quality at all latencies for both AL and LAAL, but there is an improvement in latency in ATD. This suggests that omission and truncation of information are occurring in the SIC corpus\footnote{This trend has also been reported by \citet{ko-etal-2023-tagged}.}. Thus, the LLM-SI Corpus, while reducing latency like SIC, maintains translation quality, unlike SIC, demonstrating its effectiveness as data for SI.

\paragraph{Evaluation 2: SIC-test}

\begin{figure*}[!t]
\centering
\includegraphics[width=\linewidth]{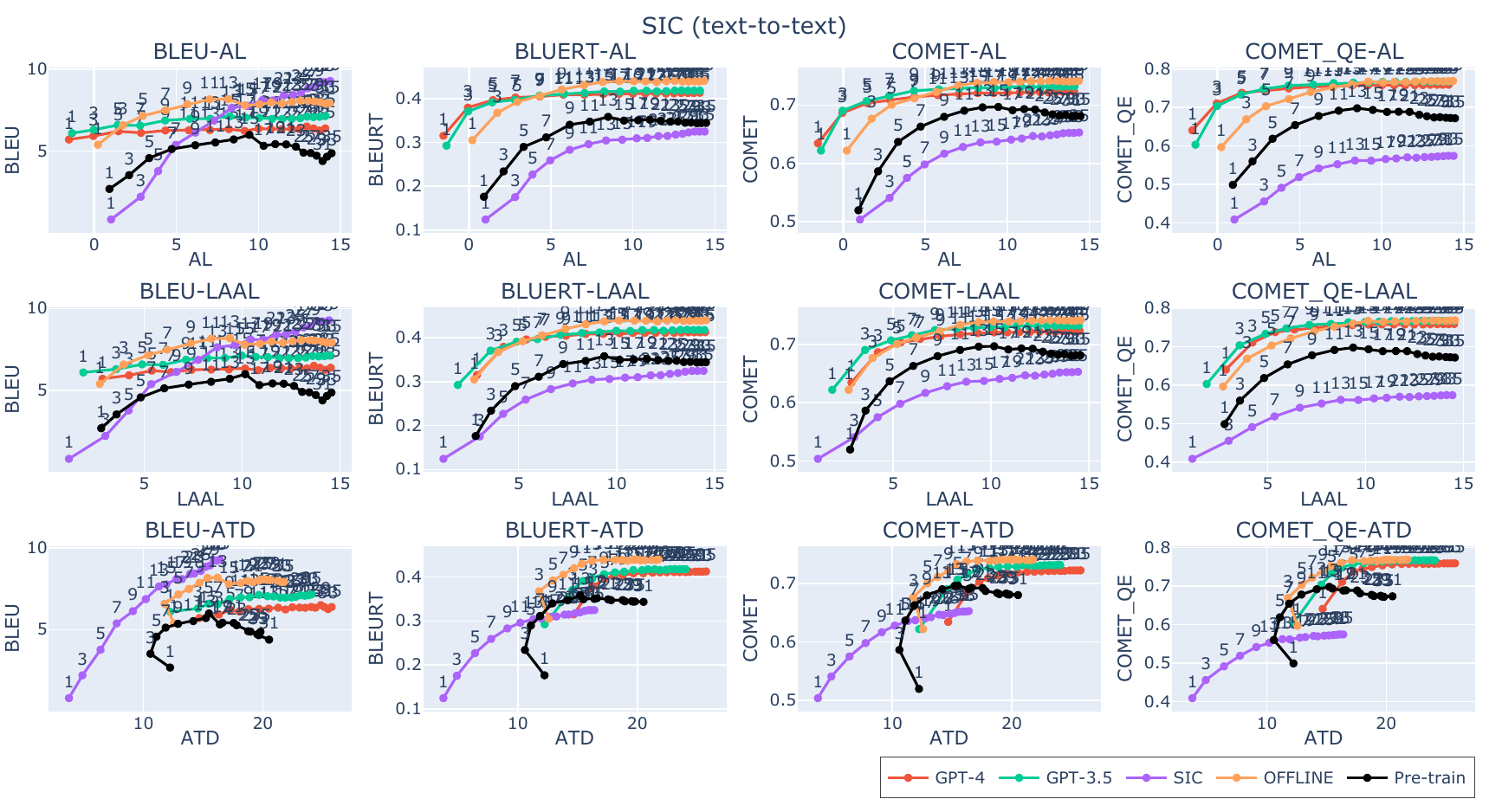}
\caption{The results of SIC-test dataset on text-to-text settings. The notations are the same as Figure \ref{fig:s2t-common}.}
\label{fig:t2t-sic}
\end{figure*}

Figure~\ref{fig:t2t-sic} shows the result of SIC-test in text-to-text settings, in which we highlight BLEU-AL, where the LLM SI-Corpus exhibits higher quality than OFFLINE up to about $k$=5. The same trend is observed in LAAL. However, SIC performs better at high latency because it aligns the training and evaluation data at the sentence level, thereby improving the BLEU score. In contrast, the LLM SI-Corpus demonstrates higher quality than SIC at low latencies. Conversely, when focusing on ATD, SIC shows the best results in both latency and quality, suggesting that the shorter output sentences are attributed to omissions and truncations. Meanwhile, when focusing on \{BLEURT, COMET, COMET-QE\}, SIC exhibits the worst translation quality. This is likely due to the effects of omissions, where missing information from the source text leads to decreased semantic similarity. Conversely, the LLM SI-Corpus outperforms OFFLINE up to a moderate level of latency, and in terms of COMET-QE, it achieves comparable or better results at all latencies.

\paragraph{Evaluation 3: Chunk-wise}

\begin{figure*}[!t]
\centering
\includegraphics[width=\linewidth]{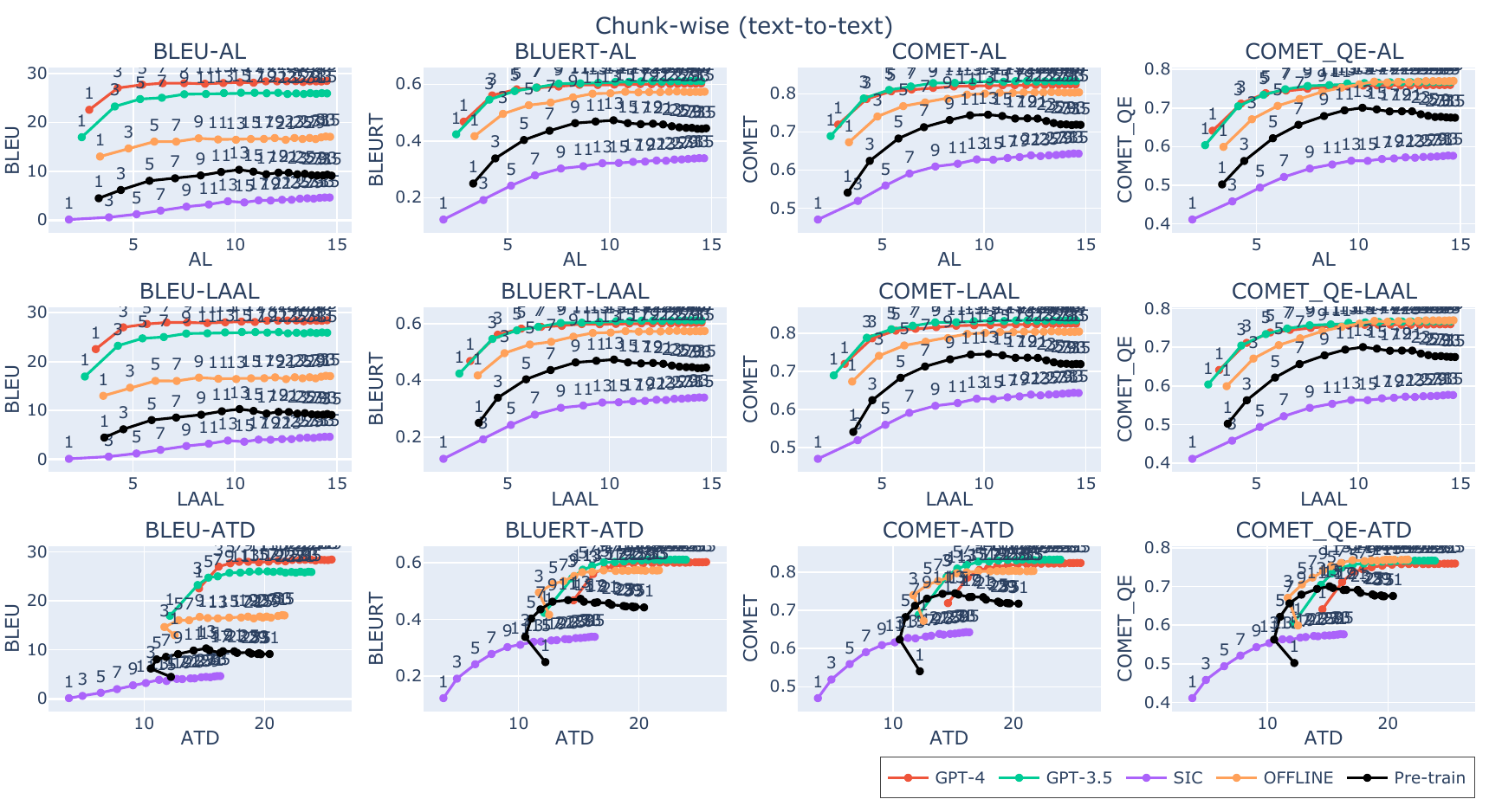}
\caption{The results of Chunk-wise dataset on text-to-text settings. The notations are the same as Figure \ref{fig:s2t-common}.}
\label{fig:t2t-chunk-wise}
\end{figure*}

Figure~\ref{fig:t2t-chunk-wise} shows the results of Chunk-wise in text-to-text settings. The LLM SI-Corpus consistently delivers the best translation qualities at all latencies. However, in ATD, although SIC has a latency advantage, its translation quality is significantly lower. Additionally, when focusing on \{AL, LAAL\}, SIC tends to translate slightly faster than the LLM SI-Corpus, suggesting that modifications such as omissions in CWMT could further improve both latency and translation quality for the LLM SI-Corpus. Finally, in the LLM SI-Corpus, there is a tendency for increased latency in ATD, indicating that longer outputs are generated.

\paragraph{Summary}

We evaluated the fine-tuned models with LLM SI-Corpus in three different test data. All results indicate that when focusing on latency, the LLM SI-Corpus delivers the best translation quality at fast latencies across all evaluation methods and datasets. Moreover, in semantic evaluation metrics using references, such as BLEURT and COMET, the LLM SI-Corpus achieves comparable or superior translation quality at all latencies. Additionally, in the reference-free metric COMET-QE, the LLM SI-Corpus consistently shows the best results in both latency and quality in all cases. However, when focusing on ATD, it is evident that the LLM SI-Corpus tends to produce longer outputs due to slightly higher latency, while it contributes to improving both latency and quality in SI models.

%%%%%%%%%%%%%%%%%%%%%%%%%%555

% \begin{table*}[!t]
% \centering
% \begin{tabular}{cp{13cm}}
% % \hline \textbf{Transcript} & \textbf{Example} \\
% \toprule
% Source & {\small So I went and met with his brother and father and said, "We're going to give you this money. What are you going to do with it?"} \\
% \midrule 
% Reference & \ja{\small お兄さんとお父さんに会い 「支援金を差し上げますが 何に使いますか？」と尋ねました} \\
% \midrule
% GPT3 & \ja{\small だから、私は彼の兄と父と会いました。そして、言いました、「わかるでしょう、このお金を渡します。」} \\
% \midrule
% GPT4 & \ja{{\small だから、私は行きました。そして、彼の兄と父親に会いました。そして、言いました、「このお金をあなたにあげますね、何をしますか?」}} \\
% \bottomrule
% \end{tabular}
% \caption{speech-to-text  waitk =7}
% \label{tab:simul_examples2}
% \end{table*}

\section{Discussions (Details)}
\label{sec:discussion-details}

\subsection{Word Order}

\begin{table*}[!t]
\centering
\begin{tabular}{cp{13cm}}
% \hline \textbf{Transcript} & \textbf{Example} \\
\toprule
Source & {\small (1) Back in New York, / (2) I am the head of development / (3) for a non-profit / (4) called Robin Hood}. \\
\midrule 
Reference & \ja{\small (1) 私は ニューヨークにある / (4) ロビンフッド財団で  / (2) 組織開発の責任者をしています。} \\
\midrule
Pretrain & \ja{\small (1) バック・イン・ニューヨーク / (2) 私は 開発部門のトップで / (4) ロビン・フッドと呼ばれます。} \\
\midrule
NAIST-SIC & \ja{\small (1)ニューヨークに戻ります。} \\
\midrule
OFFLINE & \ja{\small (1)バック・イン・ニューヨークでは 、/ (4) 私は、ロビン・フッドという / (3) 非営利団体の、/ (2) 開発部門のトップです。} \\
\midrule
GPT-3.5 & \ja{\small (1) ニューヨークに戻ると / (2) 私は開発の責任者です。/ (3) 非利益のために、/ (4) ロビンフッドと呼ばれる。}\\
\midrule
GPT-4 & \ja{{\small (1)ニューヨークに戻って、/ (2) 私はその開発の責任者です。/ (3)それは、非営利のための、/ (4) ロビンフッドと呼ばれる利益のためのものです。}} \\
\bottomrule
\end{tabular}
\caption{Example of output sentences in Pre-train, NAIST-SIC, OFFLINE, GPT-3.5, and GPT-4 on tst-COMMON in wait-$k$=7 on Text-to-Text setting.}
\label{tab:nyc_simul_examples}
\end{table*}

\begin{table*}[!t]
\centering
\begin{tabular}{cp{13cm}}
% \hline \textbf{Transcript} & \textbf{Example} \\
\toprule
Source & {\small (1) And I spent 30 days / (2) eating nothing but this -- / (3) fun in the beginning, / (4) little difficult in the middle, / (5) very dangerous in the end.} \\
\midrule 
Reference & \ja{\small (1) そしてこればかり30日間 / (2) 食べたときは / (3) 最初は楽しかったのが / (4) 途中で困難に / (5) 最後には非常に危険となりました.} \\
\midrule
Pretrain & \ja{{\small 始めにこんなことを (1) 30日も/ (2)  食べていました / (3) 楽しいことばかりです/ (4) 中間に少しは難しいのですが / (5) とても危険です}} \\
\midrule
NAIST-SIC & \ja{\small (1) 三十日 / 、(2) これ、これ、これ、これ、これ、これ、これ、これ、これ、これ、これ、これ、これ.....} \\
\midrule
OFFLINE & \ja{{\small (1) 30日も、/ (2) こんなことを、何も食べませんでした / 、(3) (笑)、最初から / 、ちょっと、ちょっと、ちょっと、ちょっと、...。}} \\
\midrule
GPT-3 & \ja{\small (3)  最初から楽しい。} \\
\midrule
GPT-4 & \ja{{\small (1)  そして、私は30日間を過ごしました/ (2) これ以外に何も食べずに、/ (3) 最初に楽しいです。 / (4) そして、中央で少し難しいです。 / (5) 最後には非常に危険です}} \\
\bottomrule
\end{tabular}
\caption{SIC fine-tuned leads to undesiable result at tst-COMMON wait-$k$ = 25 on Speech-To-Text settings.}
\label{tab:not-good-sic_examples}
\end{table*}

\begin{table*}[!t]
\centering
\begin{tabular}{cp{13cm}}
% \hline \textbf{Transcript} & \textbf{Example} \\
\toprule
Source & {\small (1) But still it was a real foot race / (2) against the other volunteers / (3) to get to the captain in charge / \underline{(4) to find out} / \underline{(5) what our assignments would be}.} \\
\midrule 
Reference & \ja{\small (3) それでも 団長を見つけて / (4) 任務を割り振ってもらうのに / (2) 他のボランティアと / (1) 激しい競走になりました。} \\
\midrule
Pretrain & \ja{\small (2) それでも足を踏みにじる他のボランティアたちに / (3) キャプテンに / (1) 足を踏みにじる真のレースでした / (5) 私たちの課題を / (4) 見つけるためです。} \\
\midrule
NAIST-SIC & \ja{\small (1)でも、} \\
\midrule
OFFLINE & \ja{\small (1) それでも、実に、アフトレースで、 / (2) 他のボランティアが / (3) キャプテンに、 / (5) 手紙を送り、 / (4) 課題を探しました 。} \\
\midrule
GPT-3.5 & \ja{\small (1) それでも、それは本物の足のレースでした 。/ (2) 他のボランティアたちに対して 、 / (3) キャプテンに向かうために 、/ \underline{(5) 私たちの課題が} / \underline{(4) 何かを見つけるために}。} \\
\midrule
GPT-4 & \ja{{\small (1) それでも、それは本当に足の運命でした。 / (2) 他のボランティアたちに対して、/ (3) キャプテンに到着するために、/ \underline{(5) 私たちの標的が何であるか} / \underline{(4) を調べるために}}} \\
\bottomrule
\end{tabular}
\caption{Example of output sentences in Pre-train, NAIST-SIC, OFFLINE, GPT-3.5, and GPT-4 on Mustc-tst-COMMON in wait$k$=7 on Text-to-Text setting. Both GPT-3.5 and GPT-4 achieve positive fluency while allowing small reordering in (4) and (5).}
\label{tab:footage_simul_examples}
\end{table*}

We investigate the extent to which word order is preserved in the source language.
Our motivation stems from the observation that in real SI scenarios, particularly in distant language pairs, interpreters endeavor to maintain word order in the source to minimize latency while preserving quality. 
To address the balance between quality and latency, we seek to emulate the guidelines employed by real SI interpreters.
Therefore, we examine how word order is preserved in the target sentences, focusing on examples generated by the wait-$k$ value of 7 (Table~\ref{tab:nyc_simul_examples}).
In the source sentences, the word order is structured as (1), (2), (3), and (4). However, in the reference sentence which comes from the subtitle of the TED Talk, the order is (1), (4), and (2), with an omission of (3). 
Both GPT-3.5 and GPT-4 fine-tuned models maintain the original word order in the source, yielding (1), (2), (3), and (4) sequences.
Conversely, OFFLINE maintains all contents from the source but rearranges them to (1), (4), (3), and (2).
In contrast, the NAIST-SIC only translates (1), omitting the rest. 
This example demonstrates that our objective of preserving word order in the source is achieved in both GPT-3.5 and GPT-4.

Both GPT-3.5 and GPT-4 achieved maintaining word order in the source.
These results suggest that while GPT-4 is considered superior to GPT-3.5 in terms of model ability, however for this task, the source language word order preservation, GPT-3.5 satisfies to fulfill the task.
%Considering the operating cost of each model, GPT-3.5 performs satisfactorily in this context.

\subsection{Quality}
We focus on the quality using reference-free metrics to avoid biases inherent in references.
Despite increasing wait-$k$ values, NAIST-SIC exhibits low output quality as observed in the outputs (Figure~\ref{fig:s2t-common}, Figure~\ref{fig:s2t-sic}, Figure~\ref{fig:s2t-chunk-wise}, Figure~\ref{fig:t2t-common}, Figure~\ref{fig:t2t-sic}, Figure~\ref{fig:t2t-common}).
Although training SiMT and SiST with real SI data is presumed beneficial for learning real-SI tactics, relying solely on SI transcripts proves inadequate for effective model training.
Similarly, Pre-trained models, characterized by fewer omissions in training data compared to the SIC dataset, still suffer from detrimental impacts on quality during testing.
OFFLINE demonstrates competitive performance on tst-COMMON, even at small wait-$k$ values such as $k$ = 3 or higher.
However, on chunk-wise datasets, these models experience quality degradation at smaller wait-$k$ values, indicating potential overfitting.
Conversely, GPT-3.5 and GPT-4 consistently deliver competitive results across both test sets.

\subsection{Latency}
In this section, our analysis regarding latency concentrates on Pre-trained, OFFLINE, GPT-3.5, and GPT-4.
We exclude NAIST-SIC due to its short outputs with poor quality in Table~\ref{tab:simul_examples_which_better}, and serious repetitions in Table~\ref{tab:not-good-sic_examples}.
In AL and LAAL, both GPT-3.5 and GPT-4 demonstrate smaller latency compared to Pre-train and OFFLINE across both text-to-text and speech-to-speech settings (Figure~\ref{fig:s2t-common}, Figure~\ref{fig:s2t-sic}, Figure~\ref{fig:s2t-chunk-wise}, Figure~\ref{fig:t2t-common}, Figure~\ref{fig:t2t-sic}, Figure~\ref{fig:t2t-common}).
In ATD, Pre-train and OFFLINE exhibit smaller latency in text-to-text settings compared to GPT-3.5 and GPT-4, whereas  LLM SI-Corpus achieves smaller latency than OFFLINE and Pre-train in speech-to-text settings.
This discrepancy arises from the tendency that Pre-trained and OFFLINE produce shorter translation outputs than GPT-3.5 and GPT-4 in text-to-text settings (Table~\ref{tab:footage_simul_examples}), serious repetitions, leading to long latency, also happen in OFFLINE in speech-to-speech settings (Table~\ref{tab:not-good-sic_examples}), and the characteristic in ATD counting both start and end timing to measure the latency.
%However,,  
%This is attributes to the serious repetition oberved in Figure~r\ref{tab:not-good-sic_examples}.

\subsection{Chunking}
\begin{figure}[!t]
\centering
\includegraphics[width=\linewidth]{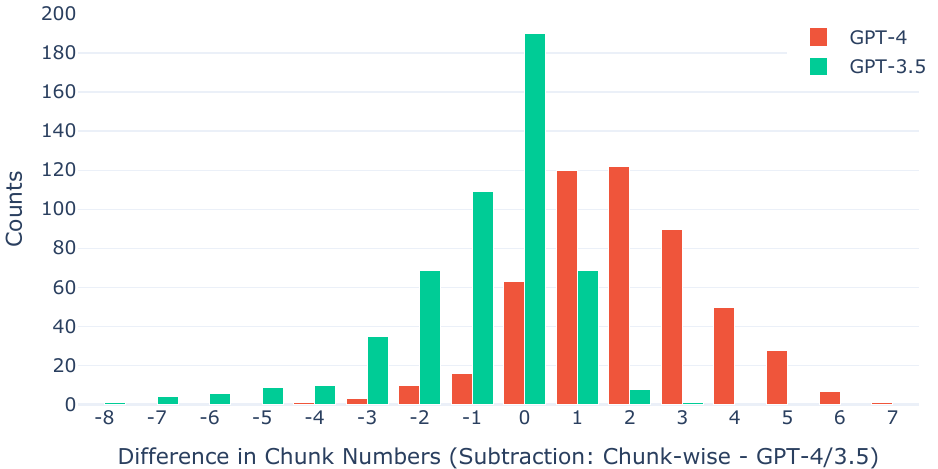}
\caption{The difference in chunk numbers between Chunk-wise and GPT-4/GPT-3.5. The total number of sentences is 511.}
\label{fig:chunk-diff}
\end{figure}

Figure~\ref{fig:chunk-diff} shows the differences in the number of chunks per sentence between Chunk-wise and LLM-SI-Corpus (GPT-3.5 and GPT-4) in the test data. These results illustrate the differences between chunking according to the chunking workflow (Chunk-wise) and chunking by the prompt (LLM-SI-Corpus). The findings in Figure~\ref{fig:chunk-diff} indicate that GPT-4 tends to chunk more finely compared to Chunk-wise, while GPT-3.5 tends to chunk more coarsely. However, chunking is merely one criterion, and the fact that these chunking results match does not necessarily mean they are good. Therefore, it is necessary to measure aspects such as word order alignment from the generated sentences, but the criteria are ambiguous, and manual measurement is not feasible. Hence, establishing automated evaluation methods is essential.

\subsection{Misalignment between Source Input and the SI data}
\label{sec:corpus-analyse}

% 文章が長すぎてうまく収まらないから後で修正する
\begin{table*}[!t]
\small
\centering
\begin{tabular}{p{7cm}p{7cm}}
% \hline \textbf{Transcript} & \textbf{Example} \\
\toprule
Source & Target \\
\midrule
\textcolor{red}{Really important.} &\ja{これが、} \\
So I'm committing to potatoes; I'm committing to milk; & \ja{{\textcolor{red}{問題なわけです。}ポテト、そしてミルク、}} \\
I'm committing to leeks and broccoli all very important stuff. & \ja{そして、ネギ、ブロッコリー、こういったものに対して、} \\
\midrule
Because of our differences, \textcolor{red}{we create and sustain life.} &\ja{違いがあるから} \\
So we should embrace our difference and aim for challenge. &\ja{\textcolor{red}{持続可能性を生み出すことができます。}} \\
\bottomrule
\end{tabular}
\caption{Example of misalignment sentence pairs in SIC.}
\label{tab:sic_misalign_examples}
\end{table*}

\begin{table*}[!t]
\small
\centering
\begin{tabular}{p{7cm}p{7cm}}
% \hline \textbf{Transcript} & \textbf{Example} \\
\toprule
Source & Target \\
\midrule
I do the philosophy of art, aesthetics, actually,\textcolor{red}{ for a living.} &\ja{私は美の哲学、美学を。} \\
I try to figure out intellectually, philosophically, and psychologically, what the experience of beauty is, what sensibly can be said about it, and how people go off the rails in trying to understand it.; & \ja{{\textcolor{red}{生業としています}、美という体験は何なのか、美について確かに言えることは何か、人は美を理解しようとして、いかに道に迷うかといったことを、知的、哲学的、心理学的に解明しようとしています。}} \\
\midrule
Now this is an extremely complicated subject, in part because the things that we call beautiful are so different. &\ja{美というのは恐ろしく込み入ったテーマであり、私たちが美しいと呼んでいるものには、非常に大きな幅があります、\textcolor{red}{いかにバラエティに富んでいることか、赤ちゃんの顔}。} \\
\red{I mean just think of the sheer variety a baby's face}, Berlioz's "Harold in Italy," movies like "The Wizard of Oz" or the plays of Chekhov, a central California landscape, a Hokusai view of Mt. Fuji, "Der Rosenkavalier," a stunning matchwinning goal in a World Cup soccer match, Van Gogh's "Starry Night," a Jane Austen novel, Fred Astaire dancing across the screen. &\ja{ベルリオーズの「イタリアのハロルド」、「オズの魔法使い」のような映画、チェーホフの戯曲、中部カリフォルニアの風景、北斎の富士山の絵、「ばらの騎士」。} \\
\bottomrule
\end{tabular}
\caption{Example of misalignment sentence pairs in Must-C.}
\label{tab:offline_misalign_examples}
\end{table*}

In our corpus analysis, we found that both NAIST-SIC-Aligned and Must-C offline data contain noise in the form of misalignment between the source and target sentences. This misalignment results in the shift of information, e.g., information in a sentence appearing in its neighbors, leading to imbalanced sentence correspondences. 
When dealing with Must-C offline data, difficulty arises in aligning audio input features with subtitles due to space limitations, which may lead to unbalanced correspondences. Similarly, in the case of NAIST-SIC-Aligned, which utilizes Japanese transcripts of interpreted data, aligning source text becomes challenging. This is due to the SI characteristics, involving omissions and summaries, which further complicate the alignment process due to imbalances between the source and target transcripts.
Some examples are shown in Table~\ref{tab:sic_misalign_examples}, Table~\ref{tab:offline_misalign_examples}. Addressing alignment in unbalanced sentences emerges as a particularly challenging aspect of SI, representing an important area for future research.

\section{Qualitative Analysis}
\label{sec:qualitative-details}

\begin{table*}[!t]
\centering
\begin{tabular}{cp{13cm}}
% \hline \textbf{Transcript} & \textbf{Example} \\
\toprule
Source & {\small \underline{(1) I  just came back from a community that} / \underline{(2) holds the secret} / \underline{(3) to human survival}}. \\
\midrule 
Reference & \ja{\small \underline{(3) 私は人類の生存に関わる} / (2) 秘密を握る / \underline{(1) あるコミュニティから戻ってきたばかりです。}} \\
\midrule
Pretrain & \ja{\small (1) ちょうどコミュニティから戻って / (2) シークレットを / (3) 人間に持つようになりました。} \\
\midrule
NAIST-SIC & \ja{\small (1) コモンティから戻って来たんです。} \\
\midrule
OFFLINE & \ja{\small (1) ちょうど、コミュニティから戻り、/ (2) シカゴに秘密を隠しました。} \\
\midrule
GPT-3.5 & \ja{\small (1) ちょうどコミュニティから戻ってきた。 / \underline{(2) それはシナリオに秘密を保持している。}/ \underline{(3) 人間の生存に。}} \\
\midrule
GPT-4 & \ja{{\small (1) ちょうど戻ってきたのは、コミュニティからで、/ \underline{(3) それは人類に} / \underline{(2) 秘密を秘めている}}。} \\
\bottomrule
\end{tabular}
\caption{Example of output sentences in Pre-train, NAIST-SIC, OFFLINE, GPT-3.5, and GPT-4 on Mustc-tst-COMMON in wait$k$=7 on Text-to-Text setting. GPT-3.5 maintains source word order completely, while GPT-4 allows small reordering in (2) and (3).}
\label{tab:community_simul_examples}
\end{table*}

\begin{table*}[!t]
\centering
\begin{tabular}{cp{13cm}}
% \hline \textbf{Transcript} & \textbf{Example} \\
\toprule
Source & {\small (1) I came to realize, / (2) as a physician, / (3) that I was working toward a goal / (4) which was different from the goal of evolution -- / \underline{(5) not necessarily contradictory, just different}.} \\
\midrule 
Reference & \ja{\small (2) 私は医師として  / (1) 気づきました / (3) 私は目標に向かって働いていますが / (4) それは進化の目標とは異なっていました / \underline{(5) 矛盾というわけではありませんが 単に違ったのです。}} \\
\midrule
Pretrain & \ja{\small (1) 気づきに来ました} \\
\midrule
NAIST-SIC & \ja{\small (1) 私が、気づいたのは、} \\
\midrule
OFFLINE & \ja{\small (1) 気づいたのです。} \\
\midrule
GPT-3.5 & \ja{\small (1) 私は気づいた。 / (2) 医師として、 / (3) 私が目指していた目標は / (4) 異なることに。} \\
\midrule
GPT-4 & \ja{{\small (1) 私は気づきました、 / (2) 医師として、 / (3) 私が取り組んでいた目標は、/ (4) それは異なる、その目標は進化の目標から、/ \underline{(5) 必ずしも矛盾するわけではない、単に異なるものです}。}} \\
\bottomrule
\end{tabular}
\caption{Example of output sentences in Pre-train, NAIST-SIC, OFFLINE, GPT-3.5, and GPT-4 on Mustc-tst-COMMON in wait$k$=7 on Text-to-Text setting. (5) is omitted in GPT-3.5, while every information is kept in GPT-4.0. }
\label{tab:realize_simul_examples}
\end{table*}

\subsection{Text-to-Text setting on tst-COMMON when $k$=7}

\begin{table*}[!t]
\centering
\begin{tabular}{cp{13cm}}
% \hline \textbf{Transcript} & \textbf{Example} \\
\toprule
Source & {\small (1) Here was some lawyer or money manager who, / (2) for the rest of his life, / \underline{(3) gets to tell people} / \underline{(4) that he went into a burning building} / \underline{(5) to save a living creature}, / \underline{(6) just because he beat me by five seconds}.} \\
\midrule 
Reference & \ja{\small (1) 弁護士だったか資産運用者だったか / (2) 彼は後々まで言い続けるでしょう / (4) 自分は燃え盛る建物の中に入り / (5) 生き物を救ったのだと / (6) 私より５秒前に着いただけなのに。} \\
\midrule
Pretrain & \ja{\small (1) ここには弁護士やお金持ちの誰かがいました} \\
\midrule
NAIST-SIC & \ja{\small (1) 弁護士やマネーマンが。} \\
\midrule
OFFLINE & \ja{\small (1) ここには弁護士やマネージャーがいます。} \\
\midrule
GPT-3.5 & \ja{\small (1) ここには弁護士やマネージャーがいました / (2) 残りの人生を過ごした。} \\
\midrule
GPT-4 & \ja{{\small (1) ここには、いくつかの弁護士またはマネージャーがいました 。/ (2) 彼は彼の生涯の残りの間、/ \underline{(3) 人々に伝え続けました}。/ \underline{(4) 彼が燃える建物に入ったと}、/ \underline{(5) 生きている生き物を救うために}。}} \\
\bottomrule
\end{tabular}
\caption{Example of output sentences in Pre-train, NAIST-SIC, OFFLINE, GPT-3.5, and GPT-4 on Mustc-tst-COMMON in wait$k$=7 on Speech-to-Text setting. From (3) to (6) is omitted in GPT-3.5, while most information is maintained in GPT-4.}
\label{tab:lawyer_simul_examples}
\end{table*}

\begin{table*}[!t]
\centering
\begin{tabular}{cp{13cm}}
% \hline \textbf{Transcript} & \textbf{Example} \\
\toprule
Source & {\small (1) So I went and met with his brother and father (2) and said, (3) "We're going to give you this money. What are you going to do with it?"} \\
\midrule 
Reference & \ja{\small (1) お兄さんとお父さんに会い / (3) 「支援金を差し上げますが 何に使いますか？」 / (2) と尋ねました} \\
\midrule
Pretrain & \ja{\small (1) それで私は彼の兄弟と父に会い / (2) こう言いました} \\
\midrule
NAIST-SIC & \ja{\small (1) 彼と会いました。、} \\
\midrule
OFFLINE & \ja{\small (1) 彼と会ったのは、兄と父親と、/ (2) こう言いました。} \\
\midrule
GPT-3.5 & \ja{\small (1) だから、私は彼の兄と父と会いました。/ (2) そして、言いました、/ (3) 「わかるでしょう、このお金を渡します} \\
\midrule
GPT-4 & \ja{\small (1) だから、私は行きました。そして、彼の兄と父親に会いました。/ (2) そして、言いました、/ (3) 「このお金をあなたにあげますね、何をしますか?」 } \\
\bottomrule
\end{tabular}
\caption{Example of output sentences in Pre-train, NAIST-SIC, OFFLINE, GPT-3.5, and GPT-4 on Mustc-tst-COMMON in wait$k$=7 on Speech-to-Text setting. GPT-4 is better than GPT-3.5 at fluency.}
\label{tab:ask_simul_examples}
\end{table*}

Table~\ref{tab:footage_simul_examples} demonstrates the equivalent qualities between GPT-3.5 and GPT-4 with a small reordering between (4) and (5).
%Our hypothesis regarding these outputs is that the source text chunk boundaries are clear with the presence of explicit chunking symbols such as \textit{against}, \textit{to}, and \textit{in charge to}.
Table~\ref{tab:community_simul_examples} shows fluency in GPT-4 is better than GPT-3.5, while both GPT-3 and GPT-4 achieve translating all contents in the source,.
This gap is attributed to small distance reordering between (2) and (3) in GPT-4, whereas word order in the source side is kept completely in GPT-3.5 ignoring fluency at each chunk boundary.
Although our motivation in this work is keeping word order in the source, we also consider small reorderings necessary to maintain its fluency. 
However, long-distance reordering, which appeared in reference as a complete switch between (1) and (3), is our focus and should not be allowed.
Such long-distance reordering leads to long latency because translating (3) in reference is possible only when (3) in the source is available and the rest is translated following (3).
Table~\ref{tab:realize_simul_examples} shows GPT-4 achieves both fluency and word order, however, the output becomes long. 
On the other hand in GPT-3, (5), the latter part in the source, is omitted.
\subsection{Speech-to-Text setting on tst-COMMON when $k$=7}
Table~\ref{tab:lawyer_simul_examples} shows the output quality gap between GPT-3.5 and 4.
The output length in GPT-4 is long but includes most chunks in the source maintaining the order in the source, whereas in GPT-3.5 only (1) and (2) are translated and the rest are omitted.
In Table~\ref{tab:ask_simul_examples}, both GPT-3.5 and GPT-4 could translate all information in the source but GPT-4 is better at quality and maintains its fluency. 
\subsection{Summary}
From these analyses, we report that while both GPT-3.5 and GPT-4 have the ability to follow the prompt to maintain the word order in the source, GPT-4 could manage the prompt and fluency at the same time better than GPT-3.5 (Table~\ref{tab:footage_simul_examples}, Table~\ref{tab:community_simul_examples}, Table~\ref{tab:ask_simul_examples}).
We also report that the severity of omitting information from the source is more serious in GPT-3.5 than GPT-4 (Table~\ref{tab:realize_simul_examples}, Table~\ref{tab:lawyer_simul_examples}). 
We reserve the ability gap between GPT-3.5 and GPT-4 as future works.

\end{document}